\documentclass{article}

\usepackage[preprint]{corl_2026}
\usepackage{amsmath}
\usepackage{amsfonts}
\usepackage{amssymb}
\usepackage{amsthm}
\usepackage{graphicx}
\usepackage{multirow}
\usepackage{booktabs}
\usepackage{makecell}
\usepackage{float}
\usepackage[table]{xcolor} 
\definecolor{paleorange}{rgb}{1.0, 0.9, 0.75} 
\usepackage{wrapfig}
\usepackage{enumitem}
\usepackage[T1]{fontenc}

\title{Object-Centric Conditioning for Visuomotor Flow Matching}
\hypersetup{
  pdftitle={Object-Centric Conditioning for Visuomotor Flow Matching},
  pdfauthor={Jijie Li, Xu Yang, Junhong Zou, Chunhai Zhao, Chaoyang Zhao, Zhen Lei, Xiangyu Zhu},
  pdfsubject={Preprint}
}

\author{
	\textbf{Jijie Li}$^{1,2}$ \quad
	\textbf{Xu Yang}$^{1,2}$ \quad
	\textbf{Junhong Zou}$^{1,2}$ \quad
	\textbf{Chunhai Zhao}$^{3}$ \\[0.5em]
	\textbf{Chaoyang Zhao}$^{1,3,\ast}$ \quad
	\textbf{Zhen Lei}$^{1,2,4,5}$ \quad
	\textbf{Xiangyu Zhu}$^{1,2,\ast}$ \\[0.5em]
	{\normalfont\small $^{1}$School of Artificial Intelligence, University of Chinese Academy of Sciences} \\
	{\normalfont\small $^{2}$State Key Laboratory of Multimodal Artificial Intelligence Systems, Institute of Automation} \\
	{\normalfont\small Chinese Academy of Sciences} \\
	{\normalfont\small $^{3}$Foundation Model Research Center, Institute of Automation, Chinese Academy of Sciences} \\
	{\normalfont\small $^{4}$Centre for Artificial Intelligence and Robotics, Hong Kong Institute of Science \& Innovation,} \\
	{\normalfont\small Chinese Academy of Sciences} \\
	{\normalfont\small $^{5}$Computer Science and Engineering, the Faculty of Innovation Engineering,} \\
	{\normalfont\small Macau University of Science and Technology} \\
	{\normalfont\small $^{\ast}$Corresponding authors} \\
	{\normalfont\small
	\shortstack{\texttt{\{lijijie2024, xu.yang, zoujunhong2022, chunhai.zhao,}\\
	\texttt{chaoyang.zhao, zhen.lei, xiangyu.zhu\}@ia.ac.cn}}}
}

\begin{document}
\maketitle

\begin{abstract}
Robot visuomotor policies are commonly formulated as autoregressive, diffusion-based, or more recently, flow matching models. Among them, Action-to-Action (A2A) flow matching improves inference efficiency by initializing generation from historical action priors rather than stochastic noise. However, stale historical motion patterns and entangled global visual representations can jointly reduce robustness under spatial out-of-distribution (OOD) shifts and visual distractors. In this work, we propose \textbf{SlotFlow}, an object-centric flow matching policy for robust visuomotor manipulation. SlotFlow decouples scene observations into semantic (``what'') features and lightweight image-plane spatial (``where'') cues to provide object-aware policy conditioning and current-state grounding. The semantic representation suppresses irrelevant background correlations, while the spatial cue improves adaptation to shifted object configurations. Extensive simulation and real-world experiments demonstrate improved robustness under visual distractors and severe spatial perturbations while preserving the low-step inference efficiency of A2A. Controlled initialization and perception ablations further identify object-centric grounding as a major source of the gains and show that it complements, rather than replaces, useful historical motion priors. Code will be released at \url{https://github.com/Li-Jijie/Object-Centric-Visuomotor-Flow}.
\end{abstract}
    
\section{Introduction}

Reliable robotic manipulation in real-world environments requires visuomotor policies to remain robust under object displacements, visual distractors, and changing scene configurations. Recent advances in visuomotor policy learning have evolved from autoregressive models~\cite{zhao2023learning} to diffusion-based generative policies~\cite{chi2025diffusion,ho2020denoising,song2020score}, and more recently, flow matching~\cite{lipman2022flow,black2024pi_0,intelligence2025pi_}, which improves inference efficiency by directly learning continuous transport trajectories. Building on this paradigm, Action-to-Action (A2A) Flow Matching~\cite{jia2026action} further replaces stochastic Gaussian initialization with historical action priors, enabling highly efficient history-conditioned trajectory generation while preserving strong manipulation performance.

\begin{figure}[t]
  \centering
  \includegraphics[width=\linewidth]{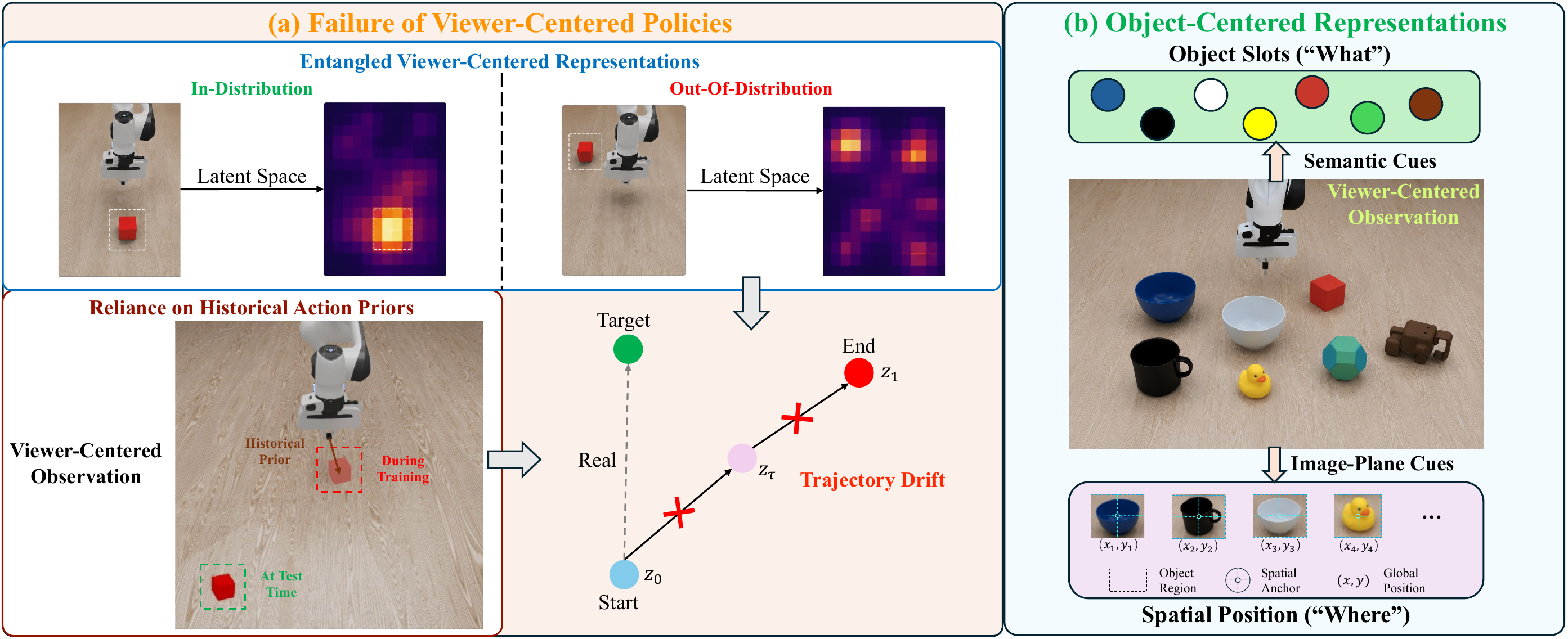}
    \caption{\textbf{Viewer-centered versus object-centric representations.}
    (a) Existing visuomotor policies rely on entangled viewer-centered representations and historical motion priors, leading to failures under spatial and visual perturbations.
    (b) Object-centric representations decouple semantic (``what'') features and image-plane spatial (``where'') cues for more robust visuomotor reasoning.}
  \label{fig:teaser}
\end{figure}

However, despite their strong in-distribution performance, history-conditioned visuomotor policies often struggle under spatial and visual perturbations commonly encountered in real-world manipulation. When object locations shift or scene appearance changes, policies may fail to sufficiently adapt generated actions to the current scene configuration. We study two coupled factors: stale historical motion patterns can conflict with a changed scene, while entangled global visual representations mix task-relevant objects with background clutter and irrelevant scene context, as illustrated in Fig.~\ref{fig:teaser}(a). Classical theories of visual perception, including Marr's object-centered representation theory~\cite{marr1982vision} and the Ungerleider and Mishkin ``what--where'' visual stream hypothesis~\cite{ungerleider1982two}, suggest that robust scene understanding benefits from transitioning from viewer-centered observations toward object-centered representations with complementary semantic (``what'') and spatial (``where'') reasoning. As illustrated in Fig.~\ref{fig:teaser}(b), object slots isolate task-relevant foreground objects from entangled scene observations, providing semantic cues, while lightweight image-plane cues encode current object locations under spatial variation.

To address these limitations, we propose \textbf{SlotFlow}, an object-centric flow matching policy for robust visuomotor manipulation. SlotFlow first extracts compact object-centric slot representations from entangled global scene observations through the proposed Cascaded Foveated Module (CFM). These object-centric slots isolate task-relevant foreground objects and decouple semantic appearance features (``what'') from lightweight image-plane spatial cues (``where''). The semantic cues provide object-aware context that suppresses irrelevant visual correlations, while the spatial cues encode the current target location for adapting flow generation to shifted object configurations. SlotFlow then injects these complementary cues into history-conditioned flow generation, strengthening current visual grounding while retaining useful temporal information. As a result, SlotFlow improves robustness under visual distractors and spatial shifts while preserving the low-step inference efficiency advantages of A2A flow matching. Our primary contributions are summarized as follows:
\begin{itemize}
    \item \textbf{Object-Centric Visual Decoupling:}
    We propose a Cascaded Foveated Module (CFM) that separates semantic features and image-plane spatial cues, effectively isolating the manipulation target from global visual distractors and scene entanglement.
    
    \item \textbf{Object-Centric Flow Conditioning:}
    We leverage semantic and spatial cues as complementary conditioning signals, providing object-aware context and current-location grounding for adapting flow generation to the current object configuration.

    \item \textbf{Robust and Efficient Flow Matching Policy:} Extensive experiments demonstrate that SlotFlow improves robustness under visual distractors and spatial out-of-distribution (OOD) shifts while preserving the low-step inference efficiency inherent to A2A flow matching.
\end{itemize}

\section{Related Work}
\label{sec:related_work}

\noindent \textbf{History-Conditioned Generative Policies.}
Visuomotor policies have evolved from deterministic and autoregressive architectures~\cite{zhao2023learning, brohan2022rt, jiang2023vima} to diffusion-based generative policies~\cite{chi2025diffusion, pearce2023imitating, Ze2024DP3, ho2020denoising, song2020score}, which improve long-horizon action generation and multimodal behavior modeling. More recently, flow matching~\cite{lipman2022flow, albergo2023building, black2024pi_0, jiang2025streaming} has emerged as an efficient alternative for continuous transport-based action generation. Building on this paradigm, history-conditioned flow policies such as A2A~\cite{jia2026action} and WarmPrior~\cite{kang2026warmprior} initialize generation from historical motion priors instead of stochastic noise to improve inference efficiency. These priors remain useful, but can become stale after scene changes; when coupled with entangled global visual representations, they may receive insufficient current-state correction under spatial OOD shifts. SlotFlow targets this coupled grounding problem through object-centric semantic features and image-plane spatial cues.

\noindent \textbf{Object-Centric Representations for Robotics and VLA.}
Object-centric representation learning~\cite{locatello2020object, burgess2019monet, greff2019multi, Zou2024TopDownGF, wu2022slotformer, kipf2021conditional} explicitly separates foreground objects from background context, demonstrating strong robustness and compositionality under dynamic scene variations. ROLA~\cite{tang2025rola} further studies object-centric representation learning for real-world scenes through attention optimization. Motivated by the need for improved visual grounding, recent robotic and VLA frameworks have explored several forms of object-centered reasoning for visuomotor control. ReconVLA~\cite{song2026reconvla}, Oat-VLA~\cite{bendikas2025focusing}, ManipLLM~\cite{li2024manipllm}, ControlVLA~\cite{li2025controlvla}, GROOT~\cite{bjorck2025gr00t}, PointACT~\cite{chen2026pointact}, and OBEYED-VLA~\cite{vo2025clutter} improve robustness through reconstruction supervision, token compression, mask conditioning, point-cloud geometry, or semantic-spatial grounding. Slot-based robotic architectures such as SlotVLA~\cite{hanyu2025slotvla}, STORM~\cite{chapin2026storm}, and SOLD~\cite{mosbach2024sold} further explore object-centric scene decomposition for robotic manipulation. In contrast, SlotFlow explicitly models semantic and spatial object-centric representations, and directly injects them into history-conditioned flow generation as complementary conditioning cues for robust visuomotor control under spatial perturbations.

\section{Method}

\begin{figure}[t]
	\centering
	\includegraphics[width=1.0\linewidth]{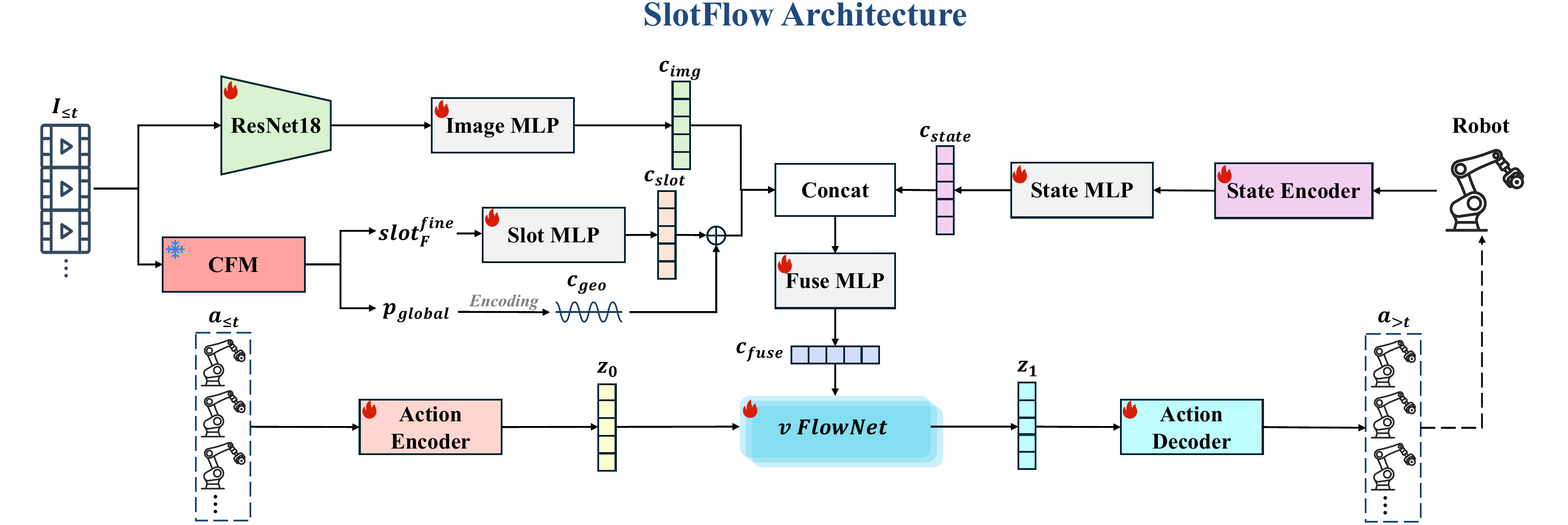}
	\caption{\textbf{SlotFlow Architecture.}
		The proposed CFM decouples observations into semantic (``what'') features and lightweight image-plane spatial (``where'') cues.}
	\label{fig:main_arch}
\end{figure}

History-conditioned flow generation requires robust grounding to current object configurations under spatial perturbations. To achieve this, we propose \textbf{SlotFlow}, an object-centric flow matching policy that explicitly decouples visual observations into semantic (``what'') and spatial (``where'') representations. As illustrated in Fig.~\ref{fig:main_arch}, SlotFlow extracts complementary object-centric representations through the proposed Cascaded Foveated Module (CFM), and injects them into history-conditioned flow generation as semantic and spatial conditioning signals. We first formulate the visuomotor policy learning problem and briefly review the preliminaries of Action-to-Action (A2A) Flow Matching~\cite{jia2026action}. We then introduce the proposed Cascaded Foveated Module (CFM), followed by the object-centric semantic and spatial conditioning mechanism for robust flow generation under spatial perturbations.

\subsection{Problem Formulation}
\label{sec:problem_formulation}

We formulate visuomotor control as learning a conditional distribution $P(A_{\text{future}} \mid o_t, A_{\text{past}})$. At each time step $t$, the policy receives an observation $o_t = \{I_t, q_t\}$, comprising an uncalibrated RGB image $I_t \in \mathbb{R}^{H \times W \times 3}$ and the robot proprioception $q_t \in \mathbb{R}^{D_q}$. Following the action chunking paradigm~\cite{zhao2023learning, chi2025diffusion}, the policy predicts a sequence of future actions $A_{\text{future}} = \{a_t, \dots, a_{t+T_p-1}\} \in \mathbb{R}^{T_p \times D_a}$ over a prediction horizon $T_p$, where $a_i$ denotes the target end-effector pose and gripper state. Furthermore, a history buffer of previously executed actions $A_{\text{past}} = \{a_{t-T_h}, \dots, a_{t-1}\} \in \mathbb{R}^{T_h \times D_a}$ is maintained to provide historical motion priors for policy generation.

\subsection{Preliminaries}
\label{sec:preliminaries}

\noindent \textbf{Flow Matching.} 
Flow matching~\cite{lipman2022flow} learns a time-dependent vector field $v_\tau$ to transform a source distribution $p_0$ into a target distribution $p_1$ via an ordinary differential equation (ODE): $d z_\tau / d \tau = v_\tau(z_\tau)$ for integration time $\tau \in [0, 1]$. Following the optimal transport path $z_\tau = (1-\tau)z_0 + \tau z_1$, a neural network $v_\theta$ is trained to approximate the vector field $v_\tau$ using the regression loss:
\begin{equation}
    \mathcal{L}_{\text{FM}} = \mathbb{E}_{\tau, z_0, z_1} \|v_\theta(z_\tau, \tau, o_t) - (z_1 - z_0)\|^2,
\label{eq:flow_matching_loss}
\end{equation}
where $z_0 \sim p_0$ represents the starting latent state, $z_1 \sim p_1$ corresponds to the target future action latents, and $o_t$ serves as the observation conditioning at the current control step $t$.

\noindent \textbf{Action-to-Action (A2A) Paradigm.} 
Unlike standard generative policies that sample the starting point from an uninformative Gaussian prior ($z_0 \sim \mathcal{N}(0, I)$), A2A~\cite{jia2026action} leverages historical motion as an informed initialization. Specifically, an encoder $E_a$ maps the historical actions $A_{\text{past}}$ into a latent prior $z_{\text{his}} = E_a(A_{\text{past}})$, which vanilla A2A directly uses as the starting point $z_0 = z_{\text{his}}$. The flow matching policy then generates future action latents conditioned on this initialized prior. This history-informed initialization significantly reduces the transport distance and enables highly efficient inference. When history becomes stale after a scene change, however, it can conflict with the current observation, and entangled global visual representations can compound this mismatch by providing insufficient current-state correction. SlotFlow therefore strengthens current-state grounding rather than assuming that the historical prior alone causes spatial-OOD failure.

\subsection{Cascaded Foveated Module (CFM)}
\label{sec:cascaded_foveated_module}

\begin{figure}[t]
	\centering
	\includegraphics[width=\linewidth]{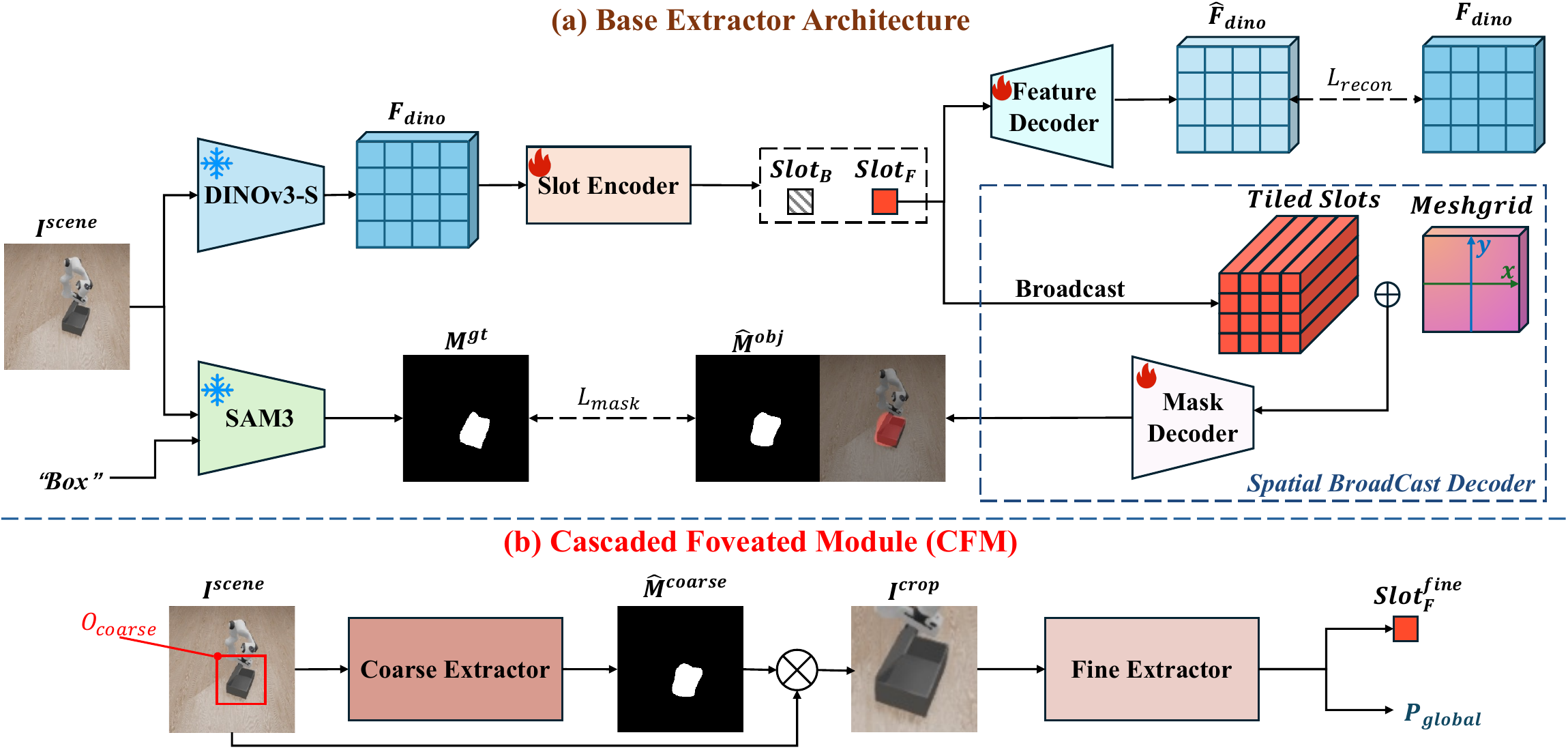}
	\caption{\textbf{CFM Overview.} 
		(a) Object-centric slot extraction with foveated feature decoding. 
		(b) Coarse-to-fine refinement of the spatial anchor $P_{\text{global}}$ and semantic representation $\text{Slot}^{\text{fine}}_F$ through localized high-resolution crops.}
	\label{fig:cfm_arch}
\end{figure}

To achieve robust object-centric perception, the proposed \textbf{Cascaded Foveated Module (CFM)} decouples observations into semantic (``what'') features and image-plane spatial (``where'') cues through a coarse-to-fine pipeline.
\paragraph{Object-Centric Slot Encoding.} CFM utilizes a frozen DINOv3-S~\cite{Simeoni2025DINOv3} and a slot encoder (Fig.~\ref{fig:cfm_arch}a) to extract background and foreground latents $\text{Slot} = \{\text{Slot}_B, \text{Slot}_F\}$. A \textit{Spatial Broadcast Decoder}~\cite{Watters2019SpatialBD} processes $\text{Slot}_F$ with a 2D meshgrid to predict an object mask $\hat{M}^{\text{obj}}$, optimized via:
\begin{equation}
    \mathcal{L}_{\text{slot}} = \alpha \mathcal{L}_{\text{mask}}(\hat{M}^{\text{obj}}, M^{\text{gt}}) + \beta \| (\hat{F}_{\text{dino}} - F_{\text{dino}}) \odot M^{\text{gt}} \|^2,
\label{eq:l_focus}
\end{equation}
where $M^{\text{gt}}$ is generated by SAM3~\cite{carion2025sam3segmentconcepts}, and $\odot$ restricts reconstruction to the foreground.
\paragraph{Cascaded Foveated Processing.}
Using a coarse-to-fine strategy (Fig.~\ref{fig:cfm_arch}b), a coarse encoder first identifies a crop region with top-left corner $O_{\text{coarse}} = (u_0, v_0)$ on the global image $I^{\text{scene}}$. The coarse stage focuses on robust semantic localization under global scene variations, while the fine stage refines local geometric grounding for precise manipulation conditioning. A high-resolution encoder then processes this crop to extract a refined semantic representation $\text{Slot}^{\text{fine}}_F$ and a fine mask $\hat{M}^{\text{fine}}$. The spatial anchor is the centroid of this fine mask:
\begin{equation}
    P_{\text{local}} = \frac{\sum_{u,v} [u,v]^T \cdot \hat{M}^{\text{fine}}(u,v)}{\sum_{u,v} \hat{M}^{\text{fine}}(u,v)}, \quad P_{\text{global}} = O_{\text{coarse}} + s \cdot P_{\text{local}},
\end{equation}
where $s$ is the crop scale factor. This allows $\text{Slot}^{\text{fine}}_F$ to capture refined object-centric semantic identity, while $P_{\text{global}}$ provides complementary spatial grounding cues for downstream flow conditioning.

\subsection{Slot-Guided Flow Matching}
\label{sec:slot_guided_fm}

In this section, we describe how the decoupled ``what'' and ``where'' representations produced by CFM are integrated into history-conditioned flow generation. Specifically, the semantic representation provides object-aware conditioning, while the geometric representation supplies complementary spatial grounding cues for robust flow generation under spatial perturbations.

\paragraph{Image-Plane Spatial Encoding.}
To incorporate spatial grounding into policy conditioning, the global spatial anchor $P_{\text{global}}$ is mapped into a learnable geometric representation $c_{\text{geo}}$ through a lightweight positional encoding function $\gamma(\cdot)$:
\begin{equation}
    c_{\text{geo}} = \gamma(P_{\text{global}}).
\end{equation}
This geometric representation provides object-centric spatial grounding cues while preserving the semantic information captured by the object-centric slot features.

\paragraph{Slot-Guided Policy Conditioning.}

As illustrated in Fig.~\ref{fig:main_arch}, the flow network is conditioned on a fused representation $c_{\text{fuse}}$ to improve object-centric grounding during flow generation. The semantic slot representation $\text{Slot}^{\text{fine}}_F$ is first projected into an object-centric semantic condition:
\begin{equation}
    c_{\text{slot}} = \text{SlotMLP}(\text{Slot}^{\text{fine}}_F).
\end{equation}

To preserve global situational awareness, the semantic condition $c_{\text{slot}}$, geometric representation $c_{\text{geo}}$, global visual features $c_{\text{img}}$, and proprioceptive state features $c_{\text{state}}$ are fused to form the final conditioning:
\begin{equation}
    c_{\text{fuse}} =
    \text{FuseMLP}
    \left(
    [c_{\text{img}}, c_{\text{state}}, c_{\text{slot}} + c_{\text{geo}}]
    \right),
\label{eq:gated_fusion}
\end{equation}
where $[\cdot]$ denotes concatenation.

This object-centric conditioning preserves global scene awareness while grounding policy generation in the target object's semantic identity and spatial configuration, thereby improving robustness under visual distractors and spatial perturbations.

\paragraph{Training Strategy.}
The proposed SlotFlow framework is trained end-to-end using the standard flow matching objective in Eq.~\ref{eq:flow_matching_loss}. During training, the CFM extracts semantic and geometric object-centric representations from input observations, which are jointly fused into the history-conditioned flow generation pathway. To improve robustness under visual perturbations, lightweight regularization and modality dropout are further applied during training. Additional implementation details are provided in the Appendix.

\section{Experiments}
\label{sec:experiments}

To evaluate SlotFlow, our empirical study is organized around three primary research questions.
\textbf{(RQ1) Efficiency \& In-Distribution Performance:}
Can SlotFlow preserve the inference efficiency and in-distribution manipulation performance of Action-to-Action (A2A) flow matching?
\textbf{(RQ2) Robustness Under Visual and Spatial Shifts:}
To what extent do object-centric semantic features and image-plane spatial cues improve robustness under visual distractors and spatial out-of-distribution (OOD) perturbations?
\textbf{(RQ3) Architectural Ablation:}
How do semantic object-centric conditioning and geometric spatial grounding individually contribute to robust visuomotor flow generation?

\subsection{Experimental Setup}
\label{sec:setup}

\noindent \textbf{Environments and Tasks.}
We evaluate SlotFlow on three Roboverse~\cite{geng2025roboverse} manipulation tasks: \textit{Close Box}, \textit{Pick Cube}, and \textit{Push Cube}. All policies operate on $448 \times 448$ RGB observations. For real-world evaluation, we further deploy SlotFlow on a UR3 robotic arm with a single RGB camera.

\noindent \textbf{OOD Evaluation Protocol.}
We evaluate robustness under three categories of out-of-distribution (OOD) perturbations:
(1) \textit{Visual Distractors:} Level 0 denotes in-distribution evaluation, while Level 1 and Level 2 introduce background replacement and illumination variation.
(2) \textit{Kinematic Perturbations:} Level 3 randomizes the robot's initial joint configuration.
(3) \textit{Spatial Generalization:} The target object is translated within the workspace plane by offsets ranging from $-0.15\text{m}$ to $0.15\text{m}$.

\noindent \textbf{Baselines and Metrics.}
We compare SlotFlow against five representative visuomotor policies: DDPM-DiT~\cite{ho2020denoising}, Score U-Net~\cite{song2020score}, FM-DiT~\cite{lipman2022flow}, VITA~\cite{gao2025vita}, and A2A~\cite{jia2026action}. We report average success over 50 rollout episodes evaluated with seeds 42/43/44. All policies use one training seed (42), so uncertainty across independently trained policies remains uncharacterized. Additional task results and controlled ablations are in the Appendix.

\begin{table}[!t]
\centering
\renewcommand{\arraystretch}{1.0}
\setlength{\tabcolsep}{10pt}
\caption{\textbf{Performance on Close Box.} Success rates under visual, kinematic, and spatial perturbations. All models are trained for 100 epochs on 99 demonstrations.}
\label{tab:main_results}
\begin{tabular}{lcccccc}
\toprule
Method & Steps & Level 0 & Level 1 & Level 2 & Level 3 & Pos Pert. \\ 
\midrule
DDPM-DiT~\cite{ho2020denoising} & 100 & 86\% & 0\% & 0\% & 0\% & 56\% \\
FM-DiT~\cite{lipman2022flow} & 10 & 98\% & 2\% & 2\% & 2\% & 60\% \\
Score-Unet~\cite{song2020score} & 100 & 84\% & 2\% & 0\% & 0\% & 58\% \\
VITA~\cite{gao2025vita} & 6 & 100\% & 4\% & 2\% & 4\% & \underline{74\%} \\
A2A~\cite{jia2026action} & 6 & 100\% & \underline{12\%} & \underline{18\%} & \underline{10\%} & 70\% \\
\midrule
\cellcolor{paleorange}\textbf{SlotFlow (Ours)} 
& \cellcolor{paleorange}6 
& \cellcolor{paleorange}\textbf{100\%} 
& \cellcolor{paleorange}\textbf{50\%} 
& \cellcolor{paleorange}\textbf{52\%} 
& \cellcolor{paleorange}\textbf{52\%} 
& \cellcolor{paleorange}\textbf{80\%} \\
\bottomrule
\end{tabular}
\end{table}

\subsection{Comparative Results}
\label{sec:comparative_results}

\begin{figure}[t]
    \centering
    \includegraphics[width=\linewidth]{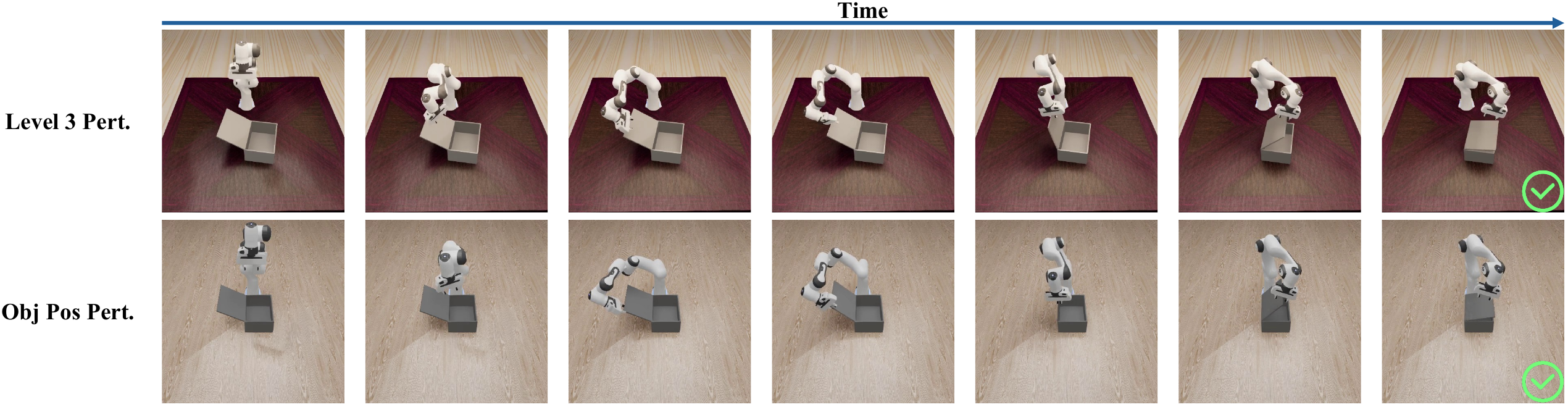} 
    \includegraphics[width=\linewidth]{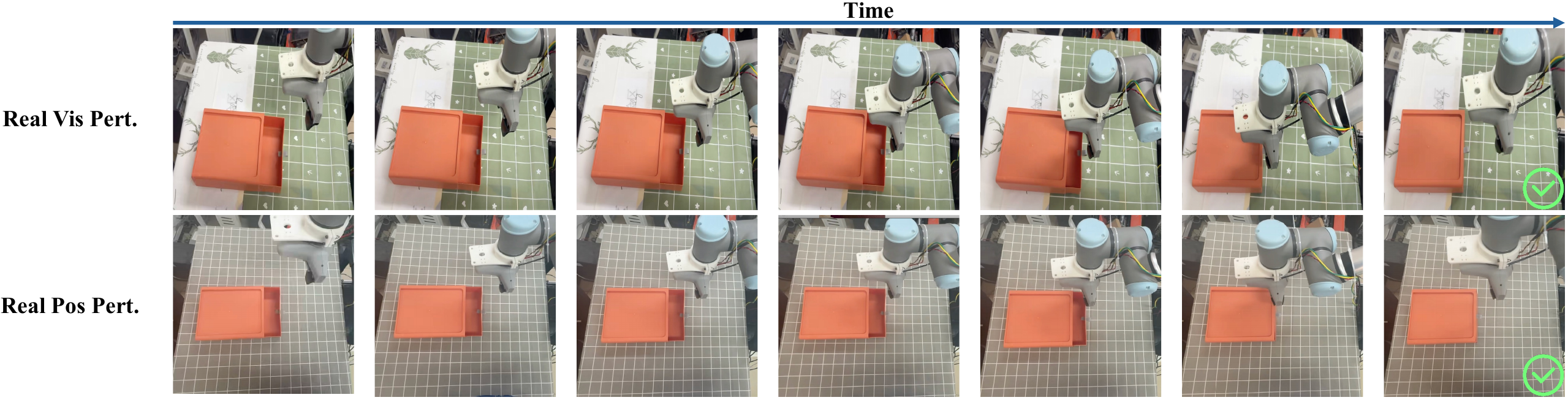}
    \caption{\textbf{Qualitative Results.} 
    Top: Simulation trajectories under Level 3 perturbations and spatial displacements. Bottom: Real-world UR3 deployment under tablecloth replacement and object position perturbations.}
    \label{fig:combined_filmstrip}
\end{figure}

Our primary evaluation analyzes policy robustness under progressive visual distractors, spatial target shifts, and initial state perturbations, as summarized in Table~\ref{tab:main_results} and qualitatively visualized in Fig.~\ref{fig:combined_filmstrip} (Top).

\noindent \textbf{Efficiency and In-Distribution Performance.}
Under Level 0 (In-Distribution), both diffusion-based and flow-matching policies achieve strong task performance, with SlotFlow reaching a 100\% success rate. Benefiting from the continuous flow matching framework, SlotFlow performs trajectory generation in only 6 inference steps, matching the efficiency of VITA and A2A while significantly reducing the sampling cost compared to standard diffusion policies such as DDPM-DiT and Score U-Net.

\noindent \textbf{Robustness Under Visual and Kinematic Perturbations.}
Level 1 to Level 3 progressively introduce background replacement, illumination variation, and initial-state perturbations. Under these conditions, standard global-feature baselines (DDPM-DiT, FM-DiT, and Score-Unet) collapse to near-zero success rates. A2A partially benefits from historical motion priors, maintaining success rates between 10\% and 18\%. In contrast, SlotFlow consistently achieves substantially higher robustness across all OOD settings, demonstrating the effectiveness of object-centric semantic conditioning under entangled visual environments.

\noindent \textbf{Spatial Generalization Under Object Translation.}
Under \textit{Pos Pert.}, the target is translated across the workspace plane, creating a mismatch between past motion and the current configuration. A2A achieves 70\%, while SlotFlow reaches 80\% by adding semantic identity and an image-plane location cue. The controlled tests below show that improved visual grounding is a major source of this gain; they do not establish the historical prior as the sole cause of A2A's errors.

\subsection{Real-World Experiments}
\label{sec:real_world}

We evaluate SlotFlow on a real-world \textit{Close Drawer} task using a UR3 robotic arm under tablecloth replacement and object position perturbations. Quantitative results are summarized in Table~\ref{tab:real_world}, while qualitative rollouts are shown in Fig.~\ref{fig:combined_filmstrip} (Bottom). Despite substantial changes in scene appearance and object configuration, SlotFlow consistently maintains stable manipulation behavior across both perturbation settings. Compared with vanilla A2A, SlotFlow achieves significantly improved robustness under both visual and positional shifts, particularly under object position perturbations where accurate geometric grounding becomes critical. These results demonstrate effective transfer of object-centric grounding from simulation to real-world robotic manipulation. Additional tasks and implementation details are provided in the Supplementary Material.

\begin{table}[htbp]
	\centering
	\renewcommand{\arraystretch}{1.0}
	\caption{\textbf{Real-World Close Drawer Performance.} Success rates under visual and spatial perturbations on a physical UR3 manipulation setup. All results are averaged over 20 real-world evaluation trials.}
	\label{tab:real_world}
	\setlength{\tabcolsep}{12pt}
	\begin{tabular}{lccc}
		\toprule
		Method & ID & Tablecloth Shift & Pos Shift \\ \midrule
		A2A~\cite{jia2026action} & 100\% & 55\% & 40\% \\
		SlotFlow (Ours) & \textbf{100\%} & \textbf{85\%} & \textbf{75\%} \\
		\bottomrule
	\end{tabular}
\end{table}

\subsection{Ablation Study}
\label{sec:ablation}

\begin{table}[!t]
\centering
\caption{\textbf{Ablation Study on Close Box.} Semantic object-centric conditioning and spatial grounding provide complementary robustness improvements across different visual backbones.}
\label{tab:ablation_results}
\setlength{\tabcolsep}{4.5pt}
\resizebox{\linewidth}{!}{%
\begin{tabular}{clcccccccc}
\toprule
Variant & Backbone & Semantic Cond. & Spatial Cond. & Level 0 & Level 1 & Level 2 & Level 3 & Pos. Pert. \\
\midrule
(a) & ResNet18 & -- & -- & 100\% & 12\% & 18\% & 10\% & 70\% \\
(b) & ResNet18 & \checkmark & -- & 100\% & 28\% & 28\% & 20\% & 66\% \\
(c) & ResNet18 & \checkmark & \checkmark & 100\% & 50\% & 52\% & 52\% & 80\% \\
\midrule
(d) & DINOv3 & -- & -- & 98\% & 18\% & 28\% & 28\% & 70\% \\
(e) & DINOv3 & \checkmark & \checkmark & 100\% & 40\% & 32\% & 34\% & 78\% \\
\bottomrule
\end{tabular}%
}
\end{table}

We evaluate five architectural variants on the \textit{Close Box} task to isolate the complementary contributions of semantic object-centric conditioning and spatial grounding (Table~\ref{tab:ablation_results}). All variants are trained under identical data and optimization settings for fair comparison (99 demonstrations and 100 epochs). Detailed component, representation, initialization, and mid-rollout controls are reported in Appendix~\ref{app:reviewer_ablations}.

\noindent \textbf{Global Representations and Visual Robustness.}
Vanilla A2A with a ResNet18 backbone suffers severe degradation under visual and spatial perturbations. Replacing ResNet18 with DINOv3 improves robustness in some settings due to stronger pretrained visual representations, but remains substantially below full SlotFlow under severe OOD shifts. Moreover, directly pooling SAM-masked DINOv3 features performs poorly. These comparisons show that stronger or localized pretrained features alone do not reproduce the benefit of task-relevant object-centric conditioning.

\noindent \textbf{Semantic and Spatial Conditioning.}
Semantic-only conditioning improves robustness to appearance changes by suppressing distractive global correlations, but it does not explicitly encode target location. Conversely, the spatial-only centroid cue provides location information without sufficiently identifying task-relevant appearance. Its 64\% position-shift result also shows that a lightweight 2D anchor alone does not replace a strong temporal prior. Combining semantic identity and the image-plane location cue yields the strongest results, supporting their complementary roles.

\noindent \textbf{Cascaded Foveated Processing.}
The no-crop variant retains both conditioning branches but removes localized fine-stage zooming, serving as a single-stage full-image control. It improves over vanilla A2A yet remains below full SlotFlow across all reported OOD settings. Thus, localized refinement contributes useful object resolution and localization accuracy, but is not the sole source of SlotFlow's robustness.

To separate initialization from grounding, we replace historical-action initialization with Gaussian noise while keeping SlotFlow's architecture, training, and six-step inference budget fixed. It obtains 36\%/32\%/24\% on Levels 1--3 and 76\% under position shift: above A2A but below full SlotFlow. In a mid-rollout test, the object is moved only after the history buffer is populated; the SlotFlow--A2A recovery gap grows from 2 points at 5 cm to 12 points at 15 cm. Thus object-centric current-state grounding is a major source of improvement and complements, rather than replaces, the temporal prior. Appendix~\ref{app:reviewer_ablations} gives full protocols and results.

\section{Conclusion}
\label{sec:conclusion}

In this paper, we presented \textbf{SlotFlow}, an object-centric flow matching policy for robust visuomotor manipulation. SlotFlow leverages the proposed Cascaded Foveated Module (CFM) to decouple visual observations into semantic features and lightweight image-plane spatial cues, which jointly improve object-aware conditioning and current-state grounding for flow generation. By introducing these complementary signals into the conditioning pathway, SlotFlow improves robustness under visual distractors and spatial OOD perturbations while preserving the low-step inference advantage of A2A flow matching. Controlled Gaussian-initialization, component, and mid-rollout perturbation experiments show that object-centric grounding is a major source of the improvement and complements useful historical motion priors. Experiments in both simulation and real-world UR3 manipulation demonstrate the effectiveness of the approach under the studied visual, kinematic, and spatial shifts, while its task-dependent scope and current limitations motivate broader future evaluation.

\section{Limitations}
\label{sec:limitations}

While SlotFlow improves robustness under the studied perturbations, its benefit is task-dependent. It is strongest for object relocation; gains are smaller for continuous-contact pushing, where temporal continuity is already informative, and for small-object precision grasping, where a 2D centroid is too coarse. The current single-target, single-view framework does not explicitly model depth, orientation, scale, camera-viewpoint changes, or severe occlusion, and it inherits A2A's vulnerability when execution errors corrupt the history buffer. Future work should extend the spatial cue to 3D and dynamic multi-object reasoning~\cite{wu2022slotformer} and validate independently trained policies across broader tasks.

\acknowledgments{This work was supported by the Strategic Priority Research Program of the Chinese Academy of Sciences under Grant XDA0480103, the National Natural Science Foundation of China under Grant 92570119, 62676398, the Science and Technology Development Fund of Macao under Project 0140/2024/AGJ, and the InnoHK program.}
\clearpage


\bibliography{example}  

@article{zhao2023learning,
  title={Learning fine-grained bimanual manipulation with low-cost hardware},
  author={Zhao, Tony Z and Kumar, Vikash and Levine, Sergey and Finn, Chelsea},
  journal={arXiv preprint arXiv:2304.13705},
  year={2023}
}

@article{chi2025diffusion,
  title={Diffusion policy: Visuomotor policy learning via action diffusion},
  author={Chi, Cheng and Xu, Zhenjia and Feng, Siyuan and Cousineau, Eric and Du, Yilun and Burchfiel, Benjamin and Tedrake, Russ and Song, Shuran},
  journal={The International Journal of Robotics Research},
  volume={44},
  number={10-11},
  pages={1684--1704},
  year={2025},
  publisher={Sage Publications Sage UK: London, England}
}

@article{lipman2022flow,
  title={Flow matching for generative modeling},
  author={Lipman, Yaron and Chen, Ricky TQ and Ben-Hamu, Heli and Nickel, Maximilian and Le, Matt},
  journal={arXiv preprint arXiv:2210.02747},
  year={2022}
}

@article{black2024pi_0,
  title={{$\pi_0$}: A Vision-Language-Action Flow Model for General Robot Control},
  author={Black, Kevin and Brown, Noah and Driess, Danny and Esmail, Adnan and Equi, Michael and Finn, Chelsea and Fusai, Niccolo and Groom, Lachy and Hausman, Karol and Ichter, Brian and others},
  journal={arXiv preprint arXiv:2410.24164},
  year={2024}
}

@article{intelligence2025pi_,
  title={{$\pi_{0.5}$}: a Vision-Language-Action Model with Open-World Generalization},
  author={Intelligence, Physical and Black, Kevin and Brown, Noah and Darpinian, James and Dhabalia, Karan and Driess, Danny and Esmail, Adnan and Equi, Michael and Finn, Chelsea and Fusai, Niccolo and others},
  journal={arXiv preprint arXiv:2504.16054},
  year={2025}
}

@article{ho2020denoising,
  title={Denoising diffusion probabilistic models},
  author={Ho, Jonathan and Jain, Ajay and Abbeel, Pieter},
  journal={Advances in neural information processing systems},
  volume={33},
  pages={6840--6851},
  year={2020}
}

@article{song2020score,
  title={Score-based generative modeling through stochastic differential equations},
  author={Song, Yang and Sohl-Dickstein, Jascha and Kingma, Diederik P and Kumar, Abhishek and Ermon, Stefano and Poole, Ben},
  journal={arXiv preprint arXiv:2011.13456},
  year={2020}
}

@article{jia2026action,
  title={Action-to-Action Flow Matching},
  author={Jia, Jindou and Li, Gen and Chen, Xiangyu and An, Tuo and Hu, Yuxuan and Li, Jingliang and Guo, Xinying and Yang, Jianfei},
  journal={arXiv preprint arXiv:2602.07322},
  year={2026}
}

@article{Watters2019SpatialBD,
  title={Spatial Broadcast Decoder: A Simple Architecture for Learning Disentangled Representations in VAEs},
  author={Nicholas Watters and Lo{\"i}c Matthey and Christopher P. Burgess and Alexander Lerchner},
  journal={ArXiv},
  year={2019},
  volume={abs/1901.07017}
}

@inproceedings{Simeoni2025DINOv3,
  title={DINOv3},
  author={Oriane Sim'eoni and Huy V. Vo and Maximilian Seitzer and Federico Baldassarre and Maxime Oquab and Cijo Jose and Vasil Khalidov and Marc Szafraniec and Seungeun Yi and Michael Ramamonjisoa and Francisco Massa and Daniel Haziza and Luca Wehrstedt and Jianyuan Wang and Timoth{\'e}e Darcet and Th{\'e}o Moutakanni and Leonel Sentana and Claire Roberts and Andrea Vedaldi and Jamie Tolan and John Brandt and Camille Couprie and Julien Mairal and Herv'e J'egou and Patrick Labatut and Piotr Bojanowski},
  year={2025}
}

@article{gao2025vita,
  title={VITA: Vision-to-Action Flow Matching Policy},
  author={Gao, Dechen and Zhao, Boqi and Lee, Andrew and Chuang, Ian and Zhou, Hanchu and Wang, Hang and Zhao, Zhe and Zhang, Junshan and Soltani, Iman},
  journal={arXiv preprint arXiv:2507.13231},
  year={2025}
}

@misc{carion2025sam3segmentconcepts,
      title={SAM 3: Segment Anything with Concepts},
      author={Nicolas Carion and Laura Gustafson and Yuan-Ting Hu and Shoubhik Debnath and Ronghang Hu and Didac Suris and Chaitanya Ryali and Kalyan Vasudev Alwala and Haitham Khedr and Andrew Huang and Jie Lei and Tengyu Ma and Baishan Guo and Arpit Kalla and Markus Marks and Joseph Greer and Meng Wang and Peize Sun and Roman Rädle and Triantafyllos Afouras and Effrosyni Mavroudi and Katherine Xu and Tsung-Han Wu and Yu Zhou and Liliane Momeni and Rishi Hazra and Shuangrui Ding and Sagar Vaze and Francois Porcher and Feng Li and Siyuan Li and Aishwarya Kamath and Ho Kei Cheng and Piotr Dollár and Nikhila Ravi and Kate Saenko and Pengchuan Zhang and Christoph Feichtenhofer},
      year={2025},
      eprint={2511.16719},
      archivePrefix={arXiv},
      primaryClass={cs.CV},
}

@article{brohan2022rt,
  title={RT-1: Robotics Transformer for Real-World Control at Scale},
  author={Anthony Brohan and Noah Brown and Justice Carbajal and Yevgen Chebotar and Joseph Dabis and Chelsea Finn and Keerthana Gopalakrishnan and Karol Hausman and Alexander Herzog and Jasmine Hsu and Julian Ibarz and Brian Ichter and Alex Irpan and Tomas Jackson and Sally Jesmonth and Nikhil J. Joshi and Ryan C. Julian and Dmitry Kalashnikov and Yuheng Kuang and Isabel Leal and Kuang-Huei Lee and Sergey Levine and Yao Lu and Utsav Malla and Deeksha Manjunath and Igor Mordatch and Ofir Nachum and Carolina Parada and Jodilyn Peralta and Emily Perez and Karl Pertsch and Jornell Quiambao and Kanishka Rao and Michael S. Ryoo and Grecia Salazar and Pannag R. Sanketi and Kevin Sayed and Jaspiar Singh and Sumedh Anand Sontakke and Austin Stone and Clayton Tan and Huong Tran and Vincent Vanhoucke and Steve Vega and Quan Ho Vuong and F. Xia and Ted Xiao and Peng Xu and Sichun Xu and Tianhe Yu and Brianna Zitkovich},
  journal={ArXiv},
  year={2022},
  volume={abs/2212.06817},
}

@inproceedings{jiang2023vima,
  title     = {VIMA: General Robot Manipulation with Multimodal Prompts},
  author    = {Yunfan Jiang and Agrim Gupta and Zichen Zhang and Guanzhi Wang and Yongqiang Dou and Yanjun Chen and Li Fei-Fei and Anima Anandkumar and Yuke Zhu and Linxi Fan},
  booktitle = {Fortieth International Conference on Machine Learning},
  year      = {2023}
}

@inproceedings{
pearce2023imitating,
title={Imitating Human Behaviour with Diffusion Models},
author={Tim Pearce and Tabish Rashid and Anssi Kanervisto and Dave Bignell and Mingfei Sun and Raluca Georgescu and Sergio Valcarcel Macua and Shan Zheng Tan and Ida Momennejad and Katja Hofmann and Sam Devlin},
booktitle={The Eleventh International Conference on Learning Representations },
year={2023},
}

@inproceedings{Ze2024DP3,
	title={3D Diffusion Policy: Generalizable Visuomotor Policy Learning via Simple 3D Representations},
	author={Yanjie Ze and Gu Zhang and Kangning Zhang and Chenyuan Hu and Muhan Wang and Huazhe Xu},
	booktitle={Proceedings of Robotics: Science and Systems (RSS)},
	year={2024}
}

@inproceedings{
albergo2023building,
title={Building Normalizing Flows with Stochastic Interpolants},
author={Michael Samuel Albergo and Eric Vanden-Eijnden},
booktitle={The Eleventh International Conference on Learning Representations },
year={2023},
}

@inproceedings{jiang2025streaming,
  title     = {Streaming Flow Policy: Simplifying diffusion/flow-matching policies by treating action trajectories as flow trajectories},
  author    = {Sunshine Jiang AND Xiaolin Fang AND Nicholas Roy AND Tom{\'a}s Lozano-P{\'e}rez AND Leslie Pack Kaelbling AND Siddharth Ancha},
  booktitle = {9th Annual Conference on Robot Learning, CoRL 2025},
  year      = {2025},
  address   = {Seoul, Korea},
  month     = {September},
}

@article{kang2026warmprior,
  title={WarmPrior: Straightening Flow-Matching Policies with Temporal Priors},
  author={Kang, Sinjae and Kim, Chanyoung and Wang, Kaixin and Zhao, Li and Lee, Kimin},
  journal={arXiv preprint arXiv:2605.13959},
  year={2026}
}

@article{locatello2020object,
  title={Object-centric learning with slot attention},
  author={Locatello, Francesco and Weissenborn, Dirk and Unterthiner, Thomas and Mahendran, Aravindh and Heigold, Georg and Uszkoreit, Jakob and Dosovitskiy, Alexey and Kipf, Thomas},
  journal={Advances in neural information processing systems},
  volume={33},
  pages={11525--11538},
  year={2020}
}

@article{burgess2019monet,
  title={Monet: Unsupervised scene decomposition and representation},
  author={Burgess, Christopher P and Matthey, Loic and Watters, Nicholas and Kabra, Rishabh and Higgins, Irina and Botvinick, Matt and Lerchner, Alexander},
  journal={arXiv preprint arXiv:1901.11390},
  year={2019}
}

@inproceedings{greff2019multi,
  title={Multi-object representation learning with iterative variational inference},
  author={Greff, Klaus and Kaufman, Rapha{\"e}l Lopez and Kabra, Rishabh and Watters, Nick and Burgess, Christopher and Zoran, Daniel and Matthey, Loic and Botvinick, Matthew and Lerchner, Alexander},
  booktitle={International conference on machine learning},
  pages={2424--2433},
  year={2019},
  organization={PMLR}
}

@inproceedings{wu2022slotformer,
  title={SlotFormer: Long-Term Dynamic Modeling in Object-Centric Models},
  author={Wu, Ziyi and Dvornik, Nikita and Greff, Klaus and Xi, Jiaqi and Kipf, Thomas and Garg, Animesh},
  booktitle={UAI 2022 Workshop on Causal Representation Learning},
  year={2022}
}

@article{kipf2021conditional,
  title={Conditional object-centric learning from video},
  author={Kipf, Thomas and Elsayed, Gamaleldin F and Mahendran, Aravindh and Stone, Austin and Sabour, Sara and Heigold, Georg and Jonschkowski, Rico and Dosovitskiy, Alexey and Greff, Klaus},
  journal={arXiv preprint arXiv:2111.12594},
  year={2021}
}

@article{hanyu2025slotvla,
  title={Slotvla: Towards modeling of object-relation representations in robotic manipulation},
  author={Hanyu, Taisei and Chung, Nhat and Le, Huy and Nguyen, Toan and Ikebe, Yuki and Gunderman, Anthony and Minh, Duy Nguyen Ho and Vo, Khoa and Kieu, Tung and Yamazaki, Kashu and others},
  journal={arXiv preprint arXiv:2511.06754},
  year={2025}
}

@article{chapin2026storm,
  title={STORM: Slot-based Task-aware Object-centric Representation for robotic Manipulation},
  author={Chapin, Alexandre and Dellandr{\'e}a, Emmanuel and Chen, Liming},
  journal={arXiv preprint arXiv:2601.20381},
  year={2026}
}

@article{mosbach2024sold,
  title={Sold: Reinforcement learning with slot object-centric latent dynamics},
  author={Mosbach, Malte and Ewertz, Jan Niklas and Villar-Corrales, Angel and Behnke, Sven},
  journal={arXiv preprint arXiv:2410.08822},
  year={2024}
}

@misc{geng2025roboverse,
      title={RoboVerse: Towards a Unified Platform, Dataset and Benchmark for Scalable and Generalizable Robot Learning}, 
      author={Haoran Geng and Feishi Wang and Songlin Wei and Yuyang Li and Bangjun Wang and Boshi An and Charlie Tianyue Cheng and Haozhe Lou and Peihao Li and Yen-Jen Wang and Yutong Liang and Dylan Goetting and Chaoyi Xu and Haozhe Chen and Yuxi Qian and Yiran Geng and Jiageng Mao and Weikang Wan and Mingtong Zhang and Jiangran Lyu and Siheng Zhao and Jiazhao Zhang and Jialiang Zhang and Chengyang Zhao and Haoran Lu and Yufei Ding and Ran Gong and Yuran Wang and Yuxuan Kuang and Ruihai Wu and Baoxiong Jia and Carlo Sferrazza and Hao Dong and Siyuan Huang and Yue Wang and Jitendra Malik and Pieter Abbeel},
      year={2025},
      eprint={2504.18904},
      archivePrefix={arXiv},
      primaryClass={cs.RO},
}

@inproceedings{Zou2024TopDownGF,
  title={Top-Down Guidance for Learning Object-Centric Representations},
  author={Junhong Zou and Xiangyu Zhu and Zhaoxiang Zhang and Zhen Lei},
  booktitle={International Joint Conference on Artificial Intelligence},
  year={2024},
}

@book{marr1982vision,
    author = {Marr, David},
    title = {Vision: A Computational Investigation into the Human Representation and Processing of Visual Information},
    publisher = {The MIT Press},
    year = {2010},
    month = {07},
    isbn = {9780262289610},
}

@article{ungerleider1982two,
title = {Object vision and spatial vision: two cortical pathways},
journal = {Trends in Neurosciences},
volume = {6},
pages = {414-417},
year = {1983},
issn = {0166-2236},
author = {Mortimer Mishkin and Leslie G. Ungerleider and Kathleen A. Macko},
}

@inproceedings{song2026reconvla,
  title={Reconvla: Reconstructive vision-language-action model as effective robot perceiver},
  author={Song, Wenxuan and Zhou, Ziyang and Zhao, Han and Chen, Jiayi and Ding, Pengxiang and Yan, Haodong and Huang, Yuxin and Tang, Feilong and Wang, Donglin and Li, Haoang},
  booktitle={Proceedings of the AAAI Conference on Artificial Intelligence},
  volume={40},
  number={22},
  pages={18549--18557},
  year={2026}
}

@article{bendikas2025focusing,
  title={Focusing on what matters: Object-agent-centric tokenization for vision language action models},
  author={Bendikas, Rokas and Dijkman, Daniel and Peschl, Markus and Haresh, Sanjay and Mazzaglia, Pietro},
  journal={arXiv preprint arXiv:2509.23655},
  year={2025}
}

@inproceedings{li2024manipllm,
  title={Manipllm: Embodied multimodal large language model for object-centric robotic manipulation},
  author={Li, Xiaoqi and Zhang, Mingxu and Geng, Yiran and Geng, Haoran and Long, Yuxing and Shen, Yan and Zhang, Renrui and Liu, Jiaming and Dong, Hao},
  booktitle={Proceedings of the IEEE/CVF Conference on Computer Vision and Pattern Recognition},
  pages={18061--18070},
  year={2024}
}

@article{li2025controlvla,
  title={Controlvla: Few-shot object-centric adaptation for pre-trained vision-language-action models},
  author={Li, Puhao and Wu, Yingying and Xi, Ziheng and Li, Wanlin and Huang, Yuzhe and Zhang, Zhiyuan and Chen, Yinghan and Wang, Jianan and Zhu, Song-Chun and Liu, Tengyu and others},
  journal={arXiv preprint arXiv:2506.16211},
  year={2025}
}

@article{vo2025clutter,
  title={Clutter-Resistant Vision-Language-Action Models through Object-Centric and Geometry Grounding},
  author={Vo, Khoa and Hanyu, Taisei and Ikebe, Yuki and Pham, Trong Thang and Chung, Nhat and Vu, Minh Nhat and Minh, Duy Nguyen Ho and Nguyen, Anh and Gunderman, Anthony and Rainwater, Chase and others},
  journal={arXiv preprint arXiv:2512.22519},
  year={2025}
}

@article{bjorck2025gr00t,
  title={Gr00t n1: An open foundation model for generalist humanoid robots},
  author={Bjorck, Johan and Casta{\~n}eda, Fernando and Cherniadev, Nikita and Da, Xingye and Ding, Runyu and Fan, Linxi and Fang, Yu and Fox, Dieter and Hu, Fengyuan and Huang, Spencer and others},
  journal={arXiv preprint arXiv:2503.14734},
  year={2025}
}

@article{chen2026pointact,
  title={PointACT: Vision-Language-Action Models with Multi-Scale Point-Action Interaction},
  author={Chen, Shizhe and Pacaud, Paul and Schmid, Cordelia},
  journal={arXiv preprint arXiv:2605.21414},
  year={2026}
}

@article{tang2025rola,
  title={ROLA: Real-World Object-Centric Learning with Attention Optimization},
  author={Tang, Qu and Wang, Haochen and Zhu, Xiangyu and Lei, Zhen and Zhang, Zhaoxiang},
  journal={Science China Information Sciences},
  volume={68},
  number={9},
  pages={192105},
  year={2025},
  publisher={Springer}
}

\clearpage
\appendix
\begin{center}
    {\huge \bf Supplementary Material}
\end{center}

\setcounter{section}{0}
\setcounter{equation}{0}
\setcounter{figure}{0}
\setcounter{table}{0}
\renewcommand{\thesection}{A\arabic{section}}
\renewcommand{\theequation}{A\arabic{equation}}
\renewcommand{\thefigure}{A\arabic{figure}}
\renewcommand{\thetable}{A\arabic{table}}
\renewcommand{\theHequation}{A.\arabic{equation}}
\renewcommand{\theHfigure}{A.\arabic{figure}}
\renewcommand{\theHtable}{A.\arabic{table}}

In this supplementary material, we provide extended empirical results, extended implementation details, and qualitative insights that were omitted from the main manuscript due to space constraints. 

\section{Extended Implementation Details}
\label{app:implementation}

\subsection{Cascaded Foveated Module (CFM)}

The proposed Cascaded Foveated Module (CFM) extracts object-centric semantic and spatial representations through a coarse-to-fine perception pipeline. We employ a frozen DINOv3-S backbone~\cite{Simeoni2025DINOv3} for both global scene encoding and object-centric feature extraction. Intermediate visual tokens are extracted from Layer 9 of the DINOv3 backbone.

Following prior object-centric representation learning frameworks~\cite{locatello2020object}, the slot encoder uses $N=2$ slots with slot dimensionality $D=256$. The resulting foreground slot representation is further processed by a lightweight mask decoder for object localization.

\paragraph{Coarse-to-Fine Foveated Processing.}
CFM adopts a two-stage foveated perception strategy. A coarse global encoder first processes the full-resolution RGB observation of size $448\times448$ to predict an approximate foreground region. The predicted object mask is decoded at resolution $224\times224$ and projected back to the original image space to obtain an object bounding box.

To improve robustness under spatial perturbations and imperfect segmentation, the predicted bounding box is expanded with a fixed 30-pixel margin. The crop size is constrained within $[0.2W,0.7W]$ and $[0.2H,0.7H]$ to avoid unstable crops. The resulting crop is resized to $448\times448$ and re-encoded by the fine-stage encoder to extract refined object-centric representations. When the mask decoder is unavailable, we use a centered fixed-scale crop as a fallback.

\paragraph{Mask Decoder.}
For foreground localization, CFM uses a Spatial Broadcast Decoder (SBD)~\cite{Watters2019SpatialBD} when pretrained decoder weights are available. The decoder operates on foreground slot features concatenated with 2D coordinate grids and predicts high-resolution foreground masks. When pretrained SBD weights are unavailable, we use a lightweight MLP-based mask prediction head.

\subsection{CFM Distillation Objectives}

To stabilize object-centric representation learning, CFM is trained with lightweight auxiliary supervision, including attention alignment, foreground/background separation, feature reconstruction, consistency regularization, and mask prediction losses. The overall distillation objective is:

\begin{equation}
\begin{aligned}
\mathcal{L}_{\text{distill}}
=
&w_{\text{attn}}(e)\mathcal{L}_{\text{attn}}
+w_{\text{bg}}(e)\mathcal{L}_{\text{bg}}
+\lambda_{\text{mass}}\mathcal{L}_{\text{mass}}
+\lambda_{\text{ent}}\mathcal{L}_{\text{ent}} \\
&+\lambda_{\text{feat}}\mathcal{L}_{\text{feat}}
+\lambda_{\text{cons}}\mathcal{L}_{\text{cons}}
+\lambda_{\text{md}}\mathcal{L}_{\text{mask}} .
\end{aligned}
\end{equation}

Here, $\mathcal{L}_{\text{attn}}$ encourages foreground/background slot attentions to align with supervision masks, $\mathcal{L}_{\text{bg}}$ suppresses foreground leakage into background regions, and $\mathcal{L}_{\text{mass}}$ encourages compact object-centric decomposition. $\mathcal{L}_{\text{ent}}$ stabilizes slot assignments, $\mathcal{L}_{\text{feat}}$ reconstructs foreground DINO features, $\mathcal{L}_{\text{cons}}$ enforces consistency under augmentation, and $\mathcal{L}_{\text{mask}}$ supervises foreground mask prediction with a combined BCE and Dice loss.

Attention and background supervision terms are linearly warmed up during the first training epochs:
\begin{equation}
w_{\text{attn}}(e)
=
\lambda_{\text{attn}}
\min(1,e/E_w),
\quad
w_{\text{bg}}(e)
=
\lambda_{\text{bg}}
\min(1,e/E_w),
\end{equation}
where $E_w=2$ denotes the warmup duration.

For all Close Box experiments, the loss coefficients are:
\begin{equation}
\lambda_{\text{attn}}=2.0,\;
\lambda_{\text{bg}}=0.4,\;
\lambda_{\text{mass}}=1.0,\;
\lambda_{\text{ent}}=0.01,
\end{equation}
\begin{equation}
\lambda_{\text{feat}}=0.5,\;
\lambda_{\text{cons}}=0.2,\;
\lambda_{\text{md}}=1.5.
\end{equation}

\subsection{Flow Policy Architecture}

The visuomotor policy is built upon the Action-to-Action (A2A) flow matching framework~\cite{jia2026action}. SlotFlow adopts a lightweight MLP-based flow network (SimpleFlowNet) for efficient low-step trajectory generation.

The flow network contains 4 hidden layers with hidden dimensionality 512. The policy predicts an action horizon of 8 future actions conditioned on 8 observation steps.

\paragraph{Object-Centric Conditioning Fusion.}
Object-centric semantic and spatial representations are integrated through conditional feature fusion. Specifically, global visual features, proprioceptive state embeddings, semantic slot representations, and spatial conditioning vectors are concatenated and projected through a joint fusion MLP.

The fused conditioning vector is injected into the flow network through conditional embedding layers prior to flow generation. To improve optimization stability, the semantic conditioning branch additionally employs a lightweight projection adapter and gating mechanism.

\subsection{Policy Optimization Objectives}

The final SlotFlow policy is optimized using the standard flow matching objective together with several lightweight auxiliary regularization losses for latent stabilization and reconstruction consistency.

\paragraph{Overall Policy Objective.}
The complete policy objective is:
\begin{equation}
\mathcal{L}_{\text{policy}}
=
\mathcal{L}_{\text{flow}}
+\lambda_{\text{cons}}\mathcal{L}_{\text{cons}}
+\lambda_{\text{rec}}\mathcal{L}_{\text{rec}}
+\lambda_{\text{reg}}\mathcal{L}_{\text{reg}},
\end{equation}
where $\mathcal{L}_{\text{flow}}$ denotes the standard A2A flow matching objective, $\mathcal{L}_{\text{cons}}$ stabilizes latent trajectory prediction under perturbations, $\mathcal{L}_{\text{rec}}$ supervises action reconstruction consistency, and $\mathcal{L}_{\text{reg}}$ contains lightweight conditioning and feature regularization terms.

The primary flow matching objective follows the standard A2A formulation:
\begin{equation}
\mathcal{L}_{\text{flow}}
=
\mathbb{E}
\|v_\theta(x_t,t,c)-u_t\|_2^2,
\end{equation}
where $v_\theta$ denotes the learned vector field and $u_t$ represents the target transport direction.

For all experiments, the auxiliary loss weights are:
\begin{equation}
\lambda_{\text{cons}}=1,\quad
\lambda_{\text{rec}}=0.5,\quad
\lambda_{\text{reg}}=10^{-4}.
\end{equation}

\subsection{Spatial Conditioning Mechanism}

To provide explicit object-centric spatial grounding, SlotFlow incorporates a lightweight spatial conditioning pathway based on the predicted object centroid.

\paragraph{Spatial Anchor Representation.}
The object centroid predicted by the fine-stage mask decoder is mapped back into the global image coordinate system and normalized into the range $[0,1]$. The normalized centroid coordinates are encoded using multi-frequency sinusoidal positional encoding and projected into a 512-dimensional spatial conditioning vector.

To stabilize training and prevent excessive reliance on spatial cues during early optimization, the injected spatial conditioning branch is modulated by a learnable zero-initialized scaling parameter.

\subsection{Training Details}

All experiments are optimized using AdamW with learning rate $1\times10^{-4}$, weight decay $1\times10^{-6}$, and momentum coefficients $(0.95,0.999)$. We employ a cosine learning rate scheduler with 500 warmup iterations.

Exponential Moving Average (EMA) updates are enabled throughout training for improved optimization stability.

For visual preprocessing, the global visual encoder follows ImageNet normalization conventions, while the DINOv3 branch uses the official DINO image processor for resizing and normalization consistency.

Although the default training configuration supports up to 1000 epochs, all experiments reported in the main paper are trained for 100 epochs unless otherwise specified.

\subsection{Inference Efficiency}

To evaluate the computational overhead introduced by object-centric perception, we compare the average policy inference latency on the \textit{Close Box} task using a single RTX 4090 GPU.

\begin{table}[htbp]
\centering
\caption{\textbf{Inference latency comparison.} Average policy inference time measured on the Close Box task.}
\label{tab:latency}
\setlength{\tabcolsep}{10pt}
\begin{tabular}{lcc}
\toprule
Method & Steps & Latency (ms) \\
\midrule
VITA & 6 & \textbf{4.40} \\
A2A & 6 & \underline{5.23} \\
SlotFlow (Ours) & 6 & 11.03 \\
FM-DiT & 10 & 21.95 \\
DDPM-DiT & 100 & 108.09 \\
Score-Unet & 100 & 116.85 \\
\bottomrule
\end{tabular}
\end{table}

Although SlotFlow introduces additional object-centric perception overhead compared with vanilla A2A, it remains substantially more efficient than iterative diffusion-based policy generation methods.

\paragraph{Model capacity.}
A2A contains 34.657M total/trainable parameters. SlotFlow contains 64.576M total parameters: 41.858M trainable and 22.718M frozen. The A2A policy backbone is unchanged; the additional capacity comes from object-centric perception and conditioning. All policies are trained once with seed 42. Evaluation episodes use rollout seeds 42, 43, and 44, which captures rollout variability but not variance across independently trained policies.

\section{Reviewer-Motivated Controlled Experiments}
\label{app:reviewer_ablations}

We report controlled experiments to distinguish current-state visual grounding from historical-action initialization and to isolate CFM's components. Unless noted otherwise, variants use the same observations, demonstrations, 100-epoch optimization budget, and six inference steps as full SlotFlow.

\begin{table}[htbp]
\centering
\caption{\textbf{Controlled ablations on Close Box.} Levels 1--3 correspond to the visual and kinematic perturbations defined in the main paper.}
\label{tab:controlled_ablations}
\setlength{\tabcolsep}{7pt}
\begin{tabular}{lcccc}
\toprule
Method & Level 1 & Level 2 & Level 3 & Pos. Pert. \\
\midrule
A2A & 12\% & 18\% & 10\% & 70\% \\
SlotFlow + Gaussian init. & 36\% & 32\% & 24\% & 76\% \\
Mask-pooled DINO & 4\% & 2\% & 4\% & 54\% \\
Spatial-only & 22\% & 22\% & 18\% & 64\% \\
No-crop SlotFlow & 26\% & 24\% & 30\% & 74\% \\
\textbf{Full SlotFlow} & \textbf{50\%} & \textbf{52\%} & \textbf{52\%} & \textbf{80\%} \\
\bottomrule
\end{tabular}
\end{table}

\paragraph{Initialization control.}
For ``SlotFlow + Gaussian init.,'' we replace the encoded historical-action starting point with Gaussian noise while holding the SlotFlow architecture, visual inputs, data, optimization, and step budget fixed. Its improvement over A2A shows that object-centric grounding is itself important; its gap to full SlotFlow shows that the historical prior remains useful. Accordingly, we do not attribute A2A's spatial-OOD failures solely to its initialization.

\paragraph{Perception controls.}
``Spatial-only'' disables semantic slot conditioning. ``Mask-pooled DINO'' supplies SAM-masked, pooled DINOv3 object features to the same A2A framework, testing whether masking and pretrained features alone suffice. ``No-crop SlotFlow'' disables fine-stage zooming and operates on the full image, testing the cascade against a single-stage alternative. The full model outperforms each control, supporting complementary semantic, spatial, and localized-refinement contributions.

\begin{table}[htbp]
\centering
\caption{\textbf{Recovery after mid-rollout object displacement.} The object is moved after the action-history buffer has been accumulated, leaving historical actions unchanged.}
\label{tab:stale_history}
\setlength{\tabcolsep}{12pt}
\begin{tabular}{lccc}
\toprule
Method & 5 cm & 10 cm & 15 cm \\
\midrule
A2A & 94\% & 80\% & 44\% \\
SlotFlow & 96\% & 84\% & 56\% \\
\textbf{Improvement} & \textbf{+2\%} & \textbf{+4\%} & \textbf{+12\%} \\
\bottomrule
\end{tabular}
\end{table}

\paragraph{Stale-history protocol.}
Unlike an initial-layout shift, this intervention creates a direct mismatch between the populated history buffer and the current scene. The widening recovery gap as displacement increases is evidence that stronger current-state grounding helps correct stale history. It does not establish the historical prior as the unique or primary source of spatial-OOD error.

\section{Extended Simulation Results}
\label{sec:supp_sim}

In this section, we provide additional experimental results on the remaining Roboverse manipulation tasks: \textit{Pick Cube} and \textit{Push Cube}. These tasks complement the main-paper \textit{Close Box} benchmark by evaluating SlotFlow under precision grasping and continuous-contact manipulation settings respectively.

\subsection{Pick Cube}
The \textit{Pick Cube} task primarily evaluates robustness under precise grasp localization and rapid end-effector motion. Compared with articulated manipulation tasks, successful grasping requires accurate spatial grounding under transient viewpoint changes and fine-grained object-relative alignment. Table~\ref{tab:pick_cube_results} summarizes quantitative results across progressive visual, kinematic, and spatial perturbation settings. Similar to the trends observed in the main paper, SlotFlow consistently improves robustness under spatial object shifts and severe visual distractors while preserving efficient low-step flow generation.

Compared with the substantial gains observed in the articulated \textit{Close Box} task, the improvements on \textit{Pick Cube} are more moderate. We hypothesize that precision grasping places stronger demands on fine-grained localization and contact timing, where the current slot-conditioned policy may still provide limited expressiveness for highly precise manipulation dynamics. Nevertheless, the consistent improvements under severe perturbations suggest that explicit semantic and spatial decoupling remains beneficial for robust history-conditioned flow matching policies. Further improving object-centric grounding for high-precision visuomotor control remains an important direction for future work.

\begin{table}[!t]
\centering
\caption{\textbf{Performance on Pick Cube.} Success rates across visual, kinematic, and spatial perturbation settings. All policies are trained for 100 epochs under identical training configurations.}
\label{tab:pick_cube_results}
\setlength{\tabcolsep}{10pt}
\begin{tabular}{lcccccc}
\toprule
Method & Steps & Level 0 & Level 1 & Level 2 & Level 3 & Pos Pert. \\ \midrule

DDPM-DiT~\cite{ho2020denoising} & 100 & 68\% & 2\% & 0\% & 0\% & \underline{32\%} \\

FM-DiT~\cite{lipman2022flow} & 10 & 74\% & 0\% & 0\% & 0\% & \underline{32\%} \\

Score-Unet~\cite{song2020score} & 100 & 68\% & 0\% & 0\% & 0\% & 28\% \\

VITA~\cite{gao2025vita} & 6 & \textbf{76\%} & 0\% & 0\% & 0\% & 24\% \\

A2A~\cite{jia2026action} & 6 & 64\% & \underline{12\%} & \underline{16\%} & \underline{16\%} & 26\% \\ \midrule

\cellcolor{paleorange}\textbf{SlotFlow (Ours)}
& \cellcolor{paleorange}6
& \cellcolor{paleorange}\textbf{76\%}
& \cellcolor{paleorange}\textbf{20\%}
& \cellcolor{paleorange}\textbf{26\%}
& \cellcolor{paleorange}\textbf{24\%}
& \cellcolor{paleorange}\textbf{38\%}
\\

\bottomrule
\end{tabular}
\end{table}

\subsection{Push Cube}
The \textit{Push Cube} task evaluates long-horizon continuous-contact manipulation under persistent interaction dynamics. Compared with grasping-based tasks, pushing requires stable object-centric conditioning throughout the motion trajectory and may accumulate spatial mismatch over long-horizon interactions. Table~\ref{tab:push_cube_results} presents the full quantitative evaluation under progressive perturbation settings. SlotFlow consistently improves robustness under spatial perturbations while maintaining efficient flow matching inference.

Compared with articulated manipulation tasks, the robustness gains on \textit{Push Cube} are relatively smaller under some perturbation settings. We hypothesize that continuous-contact pushing exhibits stronger temporal continuity and lower phase ambiguity, so the history-conditioned prior already captures a substantial portion of the motion regularity. In such settings, explicit object-centric re-grounding may be complementary rather than dominant. Nevertheless, SlotFlow still consistently improves robustness under severe perturbations, indicating that semantic and spatial conditioning remain useful across diverse manipulation regimes.

\begin{table}[!t]
\centering
\caption{\textbf{Performance on Push Cube.} Success rates across visual, kinematic, and spatial perturbation settings. All policies are trained for 100 epochs under identical training configurations.}
\label{tab:push_cube_results}
\setlength{\tabcolsep}{10pt}
\begin{tabular}{lcccccc}
\toprule
Method & Steps & Level 0 & Level 1 & Level 2 & Level 3 & Pos Pert. \\ \midrule

DDPM-DiT~\cite{ho2020denoising} & 100 & 90\% & 24\% & 26\% & 22\% & \underline{82\%} \\

FM-DiT~\cite{lipman2022flow} & 10 & 92\% & 0\% & 4\% & 4\% & 76\% \\

Score-Unet~\cite{song2020score} & 100 & \textbf{96\%} & 24\% & 16\% & 24\% & 80\% \\

VITA~\cite{gao2025vita} & 6 & 92\% & 18\% & 10\% & 6\% & 80\% \\

A2A~\cite{jia2026action} & 6 & 92\% & \textbf{72\%} & \underline{72\%} & \textbf{74\%} & 80\% \\ \midrule

\cellcolor{paleorange}\textbf{SlotFlow (Ours)}
& \cellcolor{paleorange}6
& \cellcolor{paleorange}\underline{94\%}
& \cellcolor{paleorange}\textbf{72\%}
& \cellcolor{paleorange}\textbf{74\%}
& \cellcolor{paleorange}\textbf{74\%}
& \cellcolor{paleorange}\textbf{86\%}
\\

\bottomrule
\end{tabular}
\end{table}

\section{Additional Real-World Evaluation}
\label{sec:supp_real_world}

To complement the cross-object generalization results presented in Section 4.3, we provide additional qualitative observations from real-world deployment on a UR3 robotic platform. The real-world evaluation focuses on spatial perturbations and visual distractors under single-camera visuomotor control. Representative rollouts will be hosted externally and linked from the project repository, including object shifts and cluttered background settings that differ from the nominal training configuration. Consistent with the simulation results, SlotFlow maintains stable manipulation behaviors under moderate spatial perturbations in real-world settings. We additionally observe that severe manipulator self-occlusion may occasionally reduce tracking stability when the target object becomes fully blocked from the camera view. This limitation suggests that more robust temporal association and occlusion-aware object grounding may further improve deployment robustness in future work.

\end{document}